\documentclass{nature}

\usepackage{array}
\usepackage[english]{babel}
\usepackage[utf8x]{inputenc}
\usepackage[colorinlistoftodos]{todonotes}

\usepackage{amsmath,amssymb,amsfonts,amsthm,bm}
\usepackage{graphicx}
\graphicspath{{./figs/}}
\usepackage{pslatex}
\usepackage{caption}
\usepackage{subcaption}
\usepackage{lineno}

\usepackage{soul}

\title{Capturing human categorization of natural images at scale \\ by combining deep networks and cognitive models}

\author{Ruairidh M. Battleday$^{1\dag*}$, $\, $Joshua C. Peterson$^{1\dag}$, $\, $and Thomas L. Griffiths$^{1}$}

\begin{document}

{\centering
\maketitle

\begin{affiliations}
 \centering 
 \item  Department of Computer Science, Princeton University
\end{affiliations}
}
\begin{center}
  $\dag$ Joint First Author.\\
  * Corresponding Author\\ruairidh.battleday@gmail.com\\$+1$ 510 990 4153\\Department of Computer Science\\ Princeton University \\ 35 Olden Street \\ Princeton, NJ, U.S.A. 08540-5233
\end{center}

\newpage
\section*{Abstract}
\begin{abstract}
Human categorization is one of the most important and successful targets of cognitive modeling in psychology, yet decades of development and assessment of competing models have been contingent on small sets of simple, artificial experimental stimuli. Here we extend this modeling paradigm to the domain of natural images, revealing the crucial role that stimulus representation plays in categorization and its implications for conclusions about how people form categories. Applying psychological models of categorization to natural images required two significant advances. First, we conducted the first large-scale experimental study of human categorization, involving over 500,000 human categorization judgments of 10,000 natural images from ten non-overlapping object categories. Second, we addressed the traditional bottleneck of representing high-dimensional images in cognitive models by exploring the best of current supervised and unsupervised deep and shallow machine learning methods. We find that selecting sufficiently expressive, data-driven representations is crucial to capturing human categorization, and using these representations allows simple models that represent categories with abstract prototypes to outperform the more complex memory-based exemplar accounts of categorization that have dominated in studies using less naturalistic stimuli. 
\end{abstract}
\section*{Introduction}

The problem of categorization---how an intelligent agent should group stimuli into discrete concepts---is an intriguing and valuable target for psychological research: it extends many influential themes in Western classical thought\cite{AristotleCats}, has clear interpretations at multiple levels of analysis\cite{anderson90}, and is  fundamental to understanding human minds and advancing artificial ones\cite{CohenHandbook}. Previous categorization research has had many successes---in particular, the development of high-precision statistical models of human behavior\cite{nosofsky98, griffiths2007unifying}. A central and enduring debate in this literature is whether human categorization is best accounted for by assuming people form abstract summary representations for categories (``prototype'' models)\cite{maddox1993comparing, reed72} or store examples of category members in memory (``exemplar'' models)\cite{mckinley1995investigations, nosofsky86}. Laboratory findings have largely favored exemplar models, which can produce more complex boundaries between categories---people can learn such boundaries when given sufficient training\cite{mckinley1995investigations,nosofsky86,nosofsky98}.

While this work has been insightful and theoretically productive, we know little about how it relates to the complex visual world it was meant to describe: it derives almost exclusively from laboratory experiments using highly-controlled and simplified perceptual stimuli, represented mathematically by hand-coded descriptions of obvious features or multidimensional-scaling (MDS) solutions of similarity judgments\cite{mckinley1995investigations, nosofsky1988similarity, juettner2000scale, palmeri2001central}. Human categorization abilities, by contrast, emerge from contact with the natural world, and the problems it poses. As the category divisions that result may be best understood in this context, a central challenge is to extend existing theory to account for behavior over such domains. Recent work has begun to take up this challenge\cite{nosofsky2017learning}; however, to truly make this extension we must first find a way to approximate psychological representations for large numbers of varied naturalistic stimuli.
%

In parallel to the efforts in psychological modeling, computer scientists have recently seen great success on the related task of natural image classification by taking the orthogonal approach of learning better feature spaces for complex naturalistic stimuli. In particular, convolutional neural networks (CNNs) have been shown to achieve human-level accuracy on several benchmark datasets by learning complex feature representations for naturalistic stimuli as the basis for their classification\cite{lecun2015deep}. In this paper, we examine whether stimulus representations from these CNNs are viable candidates for the approximation of inaccessible human ones, paving the way towards a wider exploration of human categorization behavior. Although it is unclear whether CNN classification models resemble human categorization or feature learning, two properties make them a promising source of representations for modeling human categorization behavior. The first is that these networks are trained on extremely large datasets of natural images, implying that they generalize broadly and offer a greater chance of approximating the experience an individual might have had with a particular class of natural objects. The second is that, given we cannot access human mental representations directly, these CNN representations have been shown to offer surprisingly good proxies for them in predicting visual cortex brain activity\cite{agrawal2014pixels,mur2013human}, and in psychological experiments investigating similarity judgments, which are closely related to categorization\cite{lake2015deep, peterson2018evaluating}. 

In order to evaluate models of human categorization of natural images using these representations, we collect and present a novel behavioral dataset, which we call \texttt{CIFAR10H}, that comprises over $500{,}000$ human categorizations of $10{,}000$ natural images from 10 categories from the ``test'' subset of the \texttt{CIFAR10}\cite{krizhevsky2009cifar} benchmark dataset used in computer vision (see Figures 1 and 2). As these images have been extensively explored by the machine learning community, they come with a wealth of corresponding classification models that may be used as the representational basis for psychological models.

We find that cognitive models are able to generate good predictions of human categorization decisions over natural images when using representations from two well-known CNN classifiers, but {\em not} when using standard hand-engineered feature extractors from computer vision or representations learned via a leading deep generative (unsupervised) CNN. In addition, when using the more effective representations, simple parametric prototype models outperform non-parametric exemplar models---a finding that contrasts with the dominance of exemplar models in laboratory research on categorization for the last 30 years\cite{mckinley1995investigations, nosofsky98, nosofsky86, zaki2003prototype}. Taken together, our findings suggest that learning the right feature space might be as fundamental to modeling categorization behavior as selecting the most suitable categorization model, especially when extending traditional theory to more naturalistic and complex domains.

\section*{Methods}
\subsection{Categorization models}
It seems intuitive that we categorize a novel stimulus based on its similarity to previously learned categories or memorized examples from them. This motivates a common framework for existing models: categorization as the assignment of a novel stimulus $\mathbf{y}$ to a category $C$ based on some measure of similarity $S(\mathbf{y},t)$ between feature vector $\mathbf{y}$ and those of existing category members (expressed in a summary statistic $t_C = f({\mathbf{x}: \mathbf{x} \in C})$). This allows us to fully specify a model by a summary statistic $t$, a similarity function $S$, and a function that links similarity scores for each category to the probability of selecting that category given $\mathbf{y}$.

The summary statistic $t$ describes the form of the category description that can vary under different categorization strategies. In a {\bf prototype model}\cite{reed72}, a category prototype---the average of category members---is used for comparison: $t_C$ becomes the central tendency $\mu_C$ of the members of category $C$.
In an {\bf exemplar model}\cite{nosofsky86}, all memorized category members are used: $t_C$ represents all existing members or ``exemplars'' of category $C$ and $\mathbf{y}$ is compared to all of them.

We take as our similarity function $S$ a standard exponentially-decreasing function of distance in the stimulus feature space\cite{shepard87}. We also take $S$ to be an additive function: if $t$ is a vector, $S$ becomes the summation of the similarities between $\mathbf{y}$ and each element of $t$. Finally, we use the Luce-Shepard choice rule\cite{luce59, shepard1957stimulus} to determine the likelihood of a single categorization, made over 10 categories:
\begin{equation}
p(\text{Guess\, } ``Category \, \, i\,"|\mathbf{y}) \,  = \, \frac{S(\mathbf{y},\, t_{Ci})^{\gamma}}{\, S(\mathbf{y},\, t_{C1})^{\gamma} \, + \, ...\, + \, S(\mathbf{y}, \, t_{C10})^{\gamma}} \,,
\label{eq:LS_basic}
\end{equation}
where $\gamma$ is a freely estimated response-scaling parameter.

Building on the common framework for specifying categorization models given above, we can reduce the probability of each judgment as follows:
\begin{equation}
p(\mbox{Guess\, } ``Category \, \, i\,"|\mathbf{y}) \, = \, \frac{1}{e^{ \, \gamma \, \log \left (\frac{S(\mathbf{y}, \, t_{C1})}{S(\mathbf{y} ,\, t_{Ci})}\right )} \, + \, \cdots \, + \, e^{ \, \gamma \, \log \left (\frac{S(\mathbf{y}, \, t_{C10})}{S(\mathbf{y} ,\, t_{Ci})}\right )} } \,.
\end{equation}
This defines a sigmoid function around the classification boundary, where $\gamma$ controls its slope, and therefore degree of determinism. As $\gamma\to\infty$, the function becomes deterministic, and as $\gamma\to 0$, it reduces to random responding. When formulated in this manner, the prototype model is equivalent to a multivariate Gaussian classifier,\cite{dudahs00} and the exemplar model to a {\it k}-nearest-neighbors classifier, where $k=N$, and distance weighting is applied.

To evaluate the predictions of these models against human data, we record the category label, $Ci$, that the participant gives to the stimulus ${\bf y}_i$. We then compute the log-likelihood of the $N$ human guesses under the model:
\begin{equation}
\mathcal{L} \, = \, \sum_{i=1}^{N} \, \log \, \frac {1} {e^{ \, \gamma \, \log \left (\frac{S(\mathbf{y}_i, \, t_{C1})}{S(\mathbf{y}_i ,\, t_{Ci})}\right )} \, + \, \cdots \, + \, e^{ \, \gamma \, \log \left (\frac{S(\mathbf{y}_i, \, t_{C10})}{S(\mathbf{y}_i ,\, t_{Ci})}\right )} }.
\end{equation}
Prototype and exemplar models differ in how their similarity to a category $S(\, {\bf y} ,\, t_C)$ is calculated, to which we now turn.

\textbf{Classic prototype models}
For prototype models, similarity to a category is taken to be an exponentially decreasing function of the distance between a stimulus vector ${\bf y}$ and the category prototype:
\begin{equation}
S(\, {\bf y} ,\, t_C) \, = \,e^{\, -d_C(\mathbf{y})} \,.
\label{eq:pt_sim}
\end{equation}
%
%
%
%
A comparison between two categories can be expressed as a simplified ratio:
\begin{equation}
\frac{S({\bf y}, t_{Cj})}{S({\bf y}, t_{Ci})} \, = \, \frac{e^{\, -d_{Cj}(\mathbf{y})}}{e^{\, -d_{Ci}(\mathbf{y})}} \, = \, e^{\, -[d_{Cj}(\mathbf{y}) \, - \, d_{Ci}(\mathbf{y})]}\, .
\label{eq:pt_compare}
\end{equation}
%
%
Classic prototype models use the (squared) Euclidean distance, $SQED\,({\bf y}, \, C) \, = \, ({\bf y}-\mu_{C})^T\, ({\bf y}-\mu_{C})$,
which gives the following similarity comparison between categories $j$ and $i$:
\begin{equation}
\frac{S({\bf y}, t_{Cj})}{S({\bf y}, t_{Ci})}  \, = \, e^{-[({\bf y}-\mu_{Cj})^T\, ({\bf y}-\mu_{Cj}) \, -\,({\bf y}-\mu_{Ci})^T\,  ({\bf y}-\mu_{Ci})]} \, = \, e^{-[SQED\,(\mathbf{y}, {C_j}) \, - \, SQED\,(\mathbf{y}, {C_i})]}\,.
\label{eq:c_pt_sim}
\end{equation}
\noindent When the prototypes are constructed by averaging the whole set of ground-truth category members, we call this the ``Classic Prototype'' model.

\textbf{Mahalanobis-distance prototype models}
These ``prototype'' models can more accurately be described as ``decision-bound'' models\cite{maddox1993comparing, griffiths2007unifying}, and there is a formal correspondence between such models in the psychological literature and a particular subset of multivariate-Gaussian classifiers from statistics in which the covariance of the Gaussian distribution describing each category is equal to the identity matrix\cite{dudahs00}. We can then use this correspondence to extend the prototype class to include several more variants by recognizing that the squared Euclidean distance is a special case of the Mahalanobis distance metric:  
\begin{align}
SQED\,(\mathbf{y}, \, C)& \, = \, (\mathbf{y}-\mu_C)^T \, \bm{I} \, (\mathbf{y}-\mu_C)\\
MHD\,(\mathbf{y}, \, C)& \, = \, 
(\mathbf{y}-\mu_C)^T \, \bm{\Sigma}_C^{-1} \, (\mathbf{y}-\mu_C) \label{eq:mh_distance}
\end{align}
\noindent where $\mu_C$ and $\bm{\Sigma}_C$ are the mean---or, prototype---and covariance matrix of category $C$. We can use this framework to define linear and quadratic decision-bound prototype models by using different strategies to estimate the covariance parameters for each ground-truth image category. 

If $\bm{\Sigma}_C$ is the same for all categories, then the decision boundary between competing prototypes in feature space is closest to is a hyperplane, resulting in a linear model\cite{dudahs00}. Taking the empirical mean of ground-truth category representations as prototype $\mathbf{\mu}_C$, we can define a linear Mahalanobis-distance model by learning a single diagonal covariance matrix common to all classes: our ``Mahalanobis Prototype (Linear)'' model. Here, $\bm{\Sigma}_C$ is a diagonal matrix, with its non-zero entries fitted to the diagonal common to all categories---i.e., $\bm{\Sigma}_{Ci} \, = \, \mbox{diag}(\mathbf{c})$, where $\mathbf{c}$ is a vector fitted on training set data across all categories.

If $\bm{\Sigma}_C$ is allowed to vary across categories, then this classification boundary can take more complex non-linear forms\cite{dudahs00}. Again taking the empirical mean of ground-truth category representations as prototype $\mathbf{\mu}_C$, we can define a quadratic Mahalanobis-distance model by learning a diagonal covariance matrix for each category: our ``Mahalanobis Prototype (Quadratic)'' model. Here, $\bm{\Sigma}_{Ci} $ is a diagonal matrix with its diagonal terms $\mathbf{c}i$ fitted on training set data for each category $Ci$.

\textbf{Hyperplane prototype models}
Whereas the above models estimate the category prototype locations from data and learn the variance terms, we can also define prototype models in which the variance terms are fixed and the prototype locations learned from data. If we do so, we can reformulate the classical prototype distance comparison in the following way:
\begin{align}
SQED\,(\mathbf{y}, {C_j}) \, - \, SQED\,(\mathbf{y}, {C_i}) \, &= \, ({\bf y}-\mu_{Cj})^T\, ({\bf y}-\mu_{Cj}) \, -\,({\bf y}-\mu_{Ci})^T\,  ({\bf y}-\mu_{Ci}) \\
&= \, {\bf y}^T{\bf y}-2{\bf \mu}_{Cj}^T{\bf y}-{\bf \mu}_{Cj}^T{\bf \mu}_{Cj} \, - ({\bf y}^T{\bf y}-2{\bf \mu}_{Ci}^T{\bf y}-{\bf \mu}_{Ci}^T{\bf \mu}_{Ci})\\
&= {\bf v}_{ji}^T{\bf y}\,+\,d_{ji} \, ,
\label{eq:c_pt_sim2}
\end{align}
\noindent where ${\bf v}_{ji} = 2({\bf \mu}_{Ci} - {\bf \mu}_{Cj}) $ defines an $(f-1)$-dimensional decision hyperplane orthogonal to the line connecting the two prototypes centered at its midpoint, and $d_{ji} = {\bf \mu}_{Ci}^T \, \, {\bf \mu}_{Ci} - {\bf \mu}_{Cj}^T \, \, {\bf \mu}_{Cj}$ is a bias term (representing the differences in sizes of the two prototypes) that can shift this hyperplane along the line. We call this our ``Hyperplane Prototype'' model, where we learn the hyperplane normal vectors, ${\bf v}_{ji}$, and bias terms, $d_{ji}$. 
%
%
%
 %

We also evaluated a number of other model variants in which {\it both} the prototype locations and diagonal variance terms were learned from training data. However, as these models performed no better than the best performing prototype described above for each set of representations, we omit them going forward.

\textbf{Exemplar models}
In exemplar models, the similarity between ${\bf y}$ and $t_C$ is given by a sum of the similarities between ${\bf y}$ and each known category member:
\begin{align}
S(\mathbf{y}, t_C ) & = \, \,\sum_{{\bf x} \in C}e^{-\beta \, d({\bf y}, {\bf x})^q} \, ,
\end{align}
where $q=2$ is a shape parameter, and $\beta$ is a ``specificity'' parameter\cite{nosofsky86}.
%
%
\noindent The distance between two vectors is given by the following equation:
\begin{equation}
d({\bf y}, {\bf x}) = \big[\sum_{k} \, w_k \mid \mathbf{x}_k \,-\,\mathbf{y}_k  \mid ^{r}\big]^{1/r} \, ,
\label{eq:distance}
\end{equation}
where we use $r=2$. Weights $w_i$, which are positive and sum to one, are known as ``attentional weights'', and serve to modify the importance of particular dimensions of the input. If these weights are fixed uniformly in advance---and in combination with the choice of $q=2$ above---we obtain an exemplar model that uses squared Euclidean distance to compare stimulus vectors (our ``Exemplar'' model). If instead we allow attention weights to vary, we obtain a more flexible model that uses the Minkowski distance to compare stimulus vectors (our ``Exemplar (Attention)'' model).
\subsection{Stimuli}
Our image stimuli were taken from the \texttt{CIFAR10} dataset, which comprises $60{,}000$  $32\times32$-pixel color images from 10 categories of natural objects\cite{krizhevsky2009cifar}. We collected human judgments for all $10{,}000$ images in the `test'' subset, which contains $1{,}000$ images for each of the following 10 categories: airplane, automobile, bird, cat, deer, dog, frog, horse, ship, truck (see Figure 1).

\begin{figure}
\centering
\includegraphics[width=\linewidth,keepaspectratio]{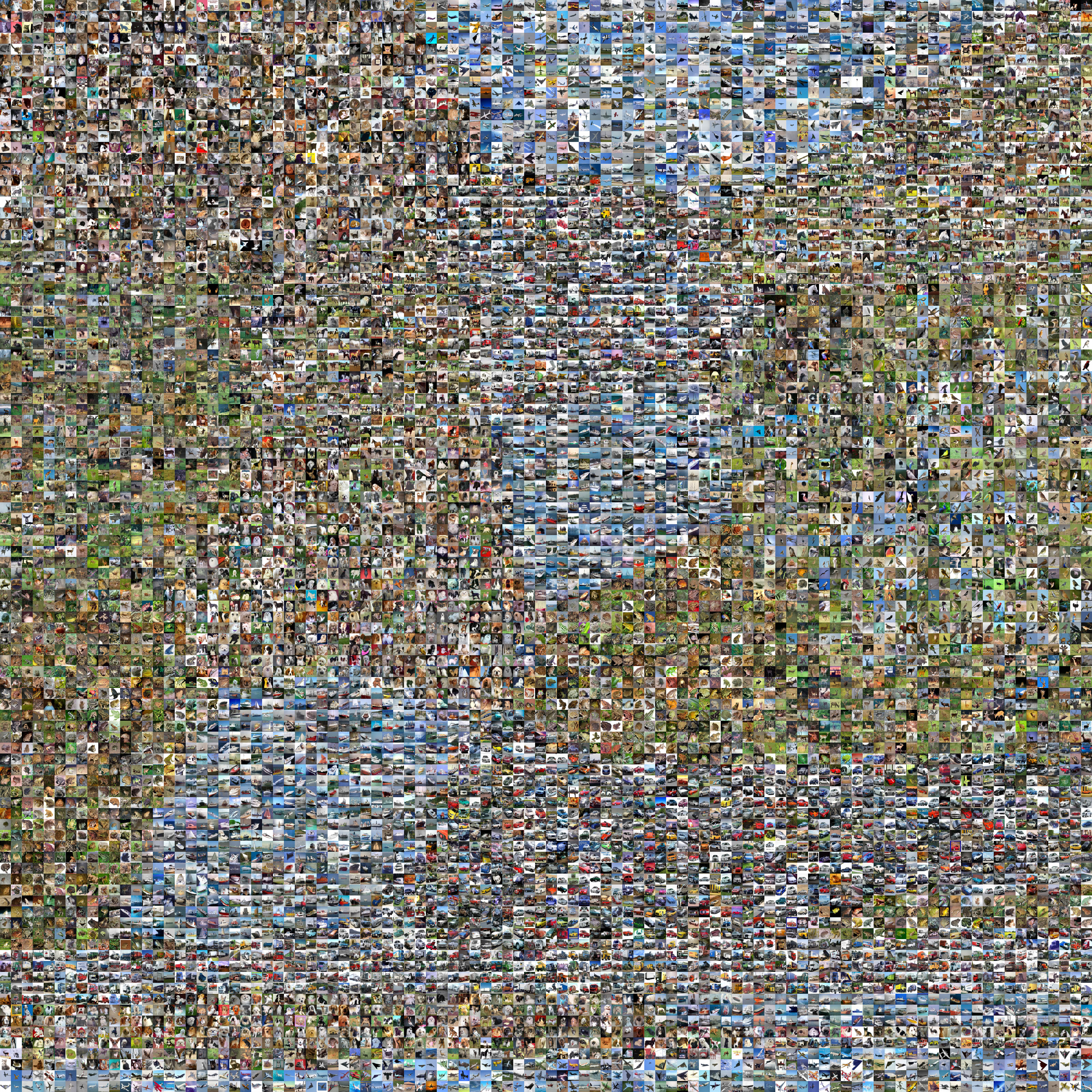}
\caption{The set of natural images we use in our experiments, organized by the similarity structure of their associated human categorizations. Images were embedded into two dimensions according to t-SNE projections of their human classification probabilities across categories\cite{maaten2008visualizing}, then mapped to a $100\!\times\!100$ grid such that adjoining images are nearest neighbors.}
\label{fig:dataset}
\end{figure}

\subsection{Human behavioral data}
Our \texttt{CIFAR10H} behavioral dataset consists of $511,400$ human categorization decisions made over our stimulus set collected via Amazon Mechanical Turk\cite{buhrmesterkg11}---to our knowledge, the largest reported in a single study to date. In our large-scale experiment, participants were shown upsampled $160\!\times\!160$-pixel images one at a time, and were asked to categorize it by clicking one of the 10 labels surrounding the image as quickly and accurately as possible (Figure 2). Label positions were shuffled between participants. There was an initial training phase, during which participants had to score at least $75\%$ accuracy, split into $3$ blocks of $20$ images taken from the \texttt{CIFAR10} training set ($60$ total, $6$ per category). If  participants failed any practice block they were asked to redo it until passing the threshold accuracy. After successful practice, each participant ($2,571$ total) categorized $200$ images ($20$ from each category) for the main experiment phase. After every $20$ trials, there was an attention check trial using a carefully selected unambiguous member of a particular category. Participants who scored below $75\%$ on these checks were removed from the final analysis ($14$ participants failed checks). The mean number of judgments per image was $51$ (range: $47 \, - \, 63$). The mean accuracy per participant was $95\%$ (range: $71\% \, - \, 100\%$). The mean accuracy per image was $95\%$ (range: $0\% \, - \, 100\%$). Average completion time was $15$ minutes, and each participant was paid $\$1.50$.

\begin{figure}
\centering
\includegraphics[width=\linewidth,keepaspectratio]{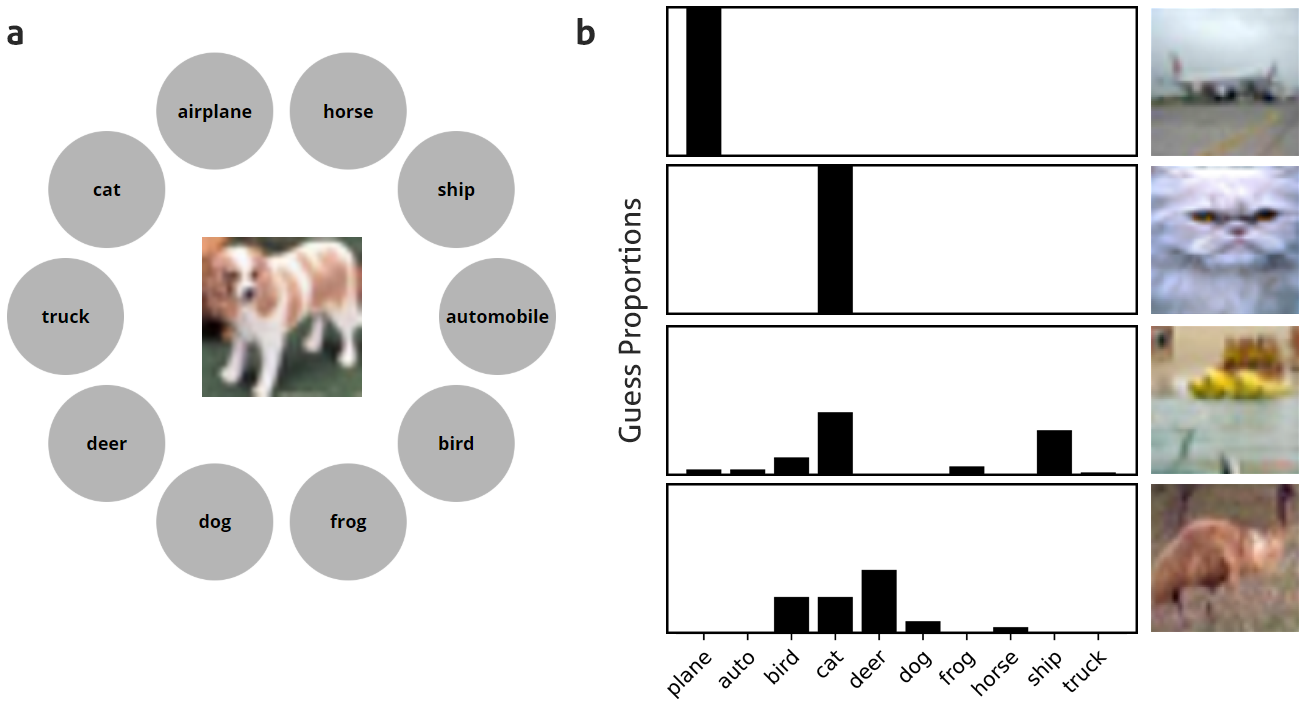}
\caption{Task paradigm and behavioral data. \textbf{a.} Experiment web interface for our human categorization task. Participants categorized each image from an order-randomized circular array of the \texttt{CIFAR10} labels. \textbf{b.} Examples of images and their human choice proportions. For many images (upper plane and cat), choices are unambiguous, matching the \texttt{CIFAR10} labels. For others (lower boat and bird), humans are far less certain.}
\label{fig:judgments}
\end{figure}

In Figure 1, we use these category judgments to help visualize our dataset. First, we take normalized judgments for each image to form a probability distribution over categories. Then, we use t-SNE projections of these probability vectors to embed the images into two dimensions\cite{maaten2008visualizing}. Finally, we place all images in a $100\!\times\!100$ grid such that adjacent images are nearest neighbors in the two-dimensional projected space. By doing this, we are able to present our images in such a way that their global and local similarity structure is preserved according to our behavioral data.

\subsection{Deep representations}
Deep CNNs learn a series of translation-invariant feature transformations of pixel-level input images that are passed to a linear classification layer in order to classify large sets of natural images\cite{lecun2015deep}. After training, a network will generate node activations at each layer for each image, forming vector representations that are increasingly abstract, and can be directly input into downstream statistical models. 

We extract feature representations for each of our stimuli from two popular off-the-shelf CNNs pre-trained on the training subset of the \texttt{CIFAR10} dataset, which comprises $5,000$ images from each of the 10 categories listed above. Our first network was a version of AlexNet for \texttt{CIFAR10}\cite{krizhevsky2012imagenet} that obtains a top-1 classification accuracy of $82\%$ on \texttt{CIFAR10}'s test subset using Caffe\cite{jia2014caffe}. We use the representations from its uppermost pooling layer, yielding $1024$-dimensional feature vectors. We use this network because it has a simple architecture that allows for easier exploration of layers while maintaining classification accuracy in the ballpark of much larger, state-of-the-art variants. For our second network, we used a DenseNet with $40$ layers, $3$ dense blocks, and a growth rate parameter of $12$, trained using the \texttt{keras} Python library to a top-1 accuracy of $93\%$\cite{chollet2015keras}. We chose this network because it achieves near state-of-the-art performance while still being parameter efficient (for faster training). The key change DenseNet makes to the fundamental CNN structure is that each layer passes its output to all layers above (i.e., ``dense'' connections), rather than just the subsequent layer, allowing for deeper networks that avoid vanishing gradients\cite{HuangDenseNet}. We use feature vectors from its uppermost pooling layer, yielding $456$-dimensional feature vectors.

We also include two further sets of image representations for comparison. The first are Histograms of Oriented Gradients (HOG) for each image\cite{lowe2004distinctive}, constructed using the Python \texttt{opencv} library\cite{opencv_library}. These features are extracted without supervision and supported numerous computer vision tasks prior to modern CNNs. Our most successful HOG representation for which we report results used a window size of $8\times8$ and $324$-dimensional feature vectors. The second set of representations are $64$-dimensional feature vectors from the latent space of a bidirectional generative adversarial network (BiGAN)\cite{dumoulin2016adversarially}. This network is convolutional, but unsupervised, and allows us to factor out the contribution of human-derived labels in training CNNs to learn representations useful for modeling people. Two-dimensional linear-discriminant-analysis plots of these representations are shown in Figure 3.

\begin{figure}
 \centering
 \includegraphics[trim={55mm 0mm 45mm 7mm},clip,width=\linewidth,keepaspectratio]{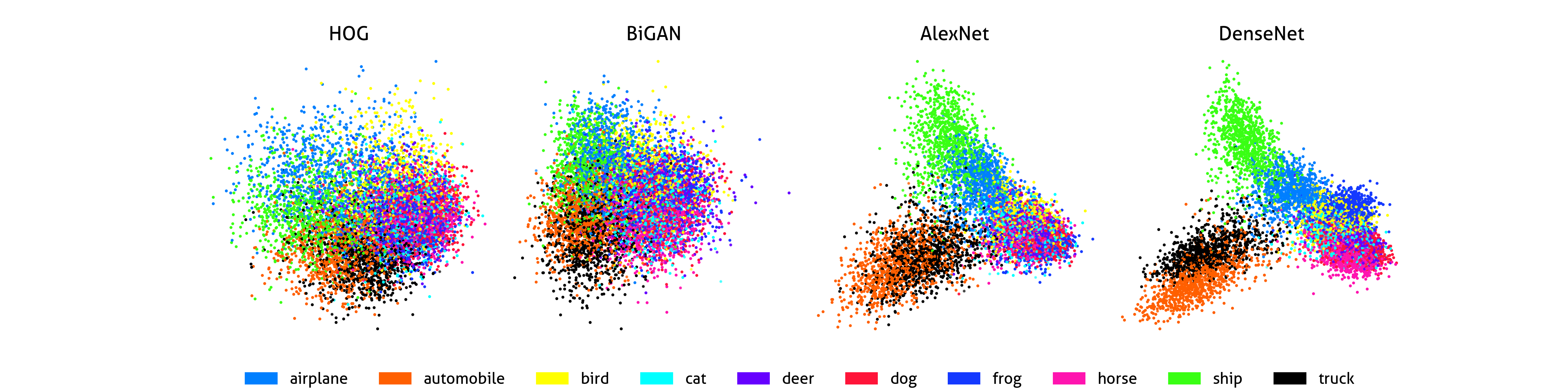}
 \caption{Two-dimensional linear-discriminant-analysis projections of the four stimulus representations we use, for all stimuli (colored by category).}
 \label{fig:LDA_plots}
\end{figure}

\subsection{Training \& Evaluation}
We optimized all model parameters with $5$-fold-cross-validation and early stopping using the adam variant of stochastic gradient descent\cite{kingmaADAM} and a batch size of $256$ images. 
For each model, we conducted a grid search over the learning and decay rate hyperparameters for adam, selecting the final model parameter set and early-stopping point during training based on which gave the lowest cross-validated average log-likelihood.

\subsection{Model comparison}
Our primary evaluation measure for each model was log-likelihood. For all models, we computed these scores by generating predictions for all images in our stimulus set using the averaged cross-validated parameters taken at the early-stopping point described above. We also use the Akaike Information Criterion (AIC) to compare models, as it gives a score for each model that takes into account the relationship between the number of parameters they employ, and their log-likelihood scores\cite{akaike1998information}
\begin{equation}
AIC \, = \, 2k \, - \, 2\ln(\hat{\mathbf{L}})
\end{equation}
where $k$ is the number of parameters in the model, and $\hat{\mathbf{L}}$ is the maximum log likelihood.

\subsection{Baselines}
As baselines, we use the raw output (``softmax'') probabilities of both AlexNet and DenseNet neural networks for each image to provide $S(\mathbf{y},\, t_C)$. In deep CNN classifiers, the softmax function takes the inner product between a matrix of learned weights and the rasterized output of the final pooling layer, and returns a probability distribution over all of the \texttt{CIFAR10H} classes. These weights are learned based on minimizing classification loss over the training subset of the \texttt{CIFAR10} dataset described above. Although not explicitly trained to output human classification probabilities, these models are the most competitive systems available in terms of making accurate classifications at the level of human performance. For this reason, and because the models output a full probability distribution over classes that may exhibit human-like confusion patterns, we expect them to provide a meaningful and competitive baseline with which to compare our model scores.

\section*{Results}
%

\begin{figure}
    \centering
	\includegraphics[trim={0mm 0mm 0mm 0mm },clip,width=\linewidth,keepaspectratio]{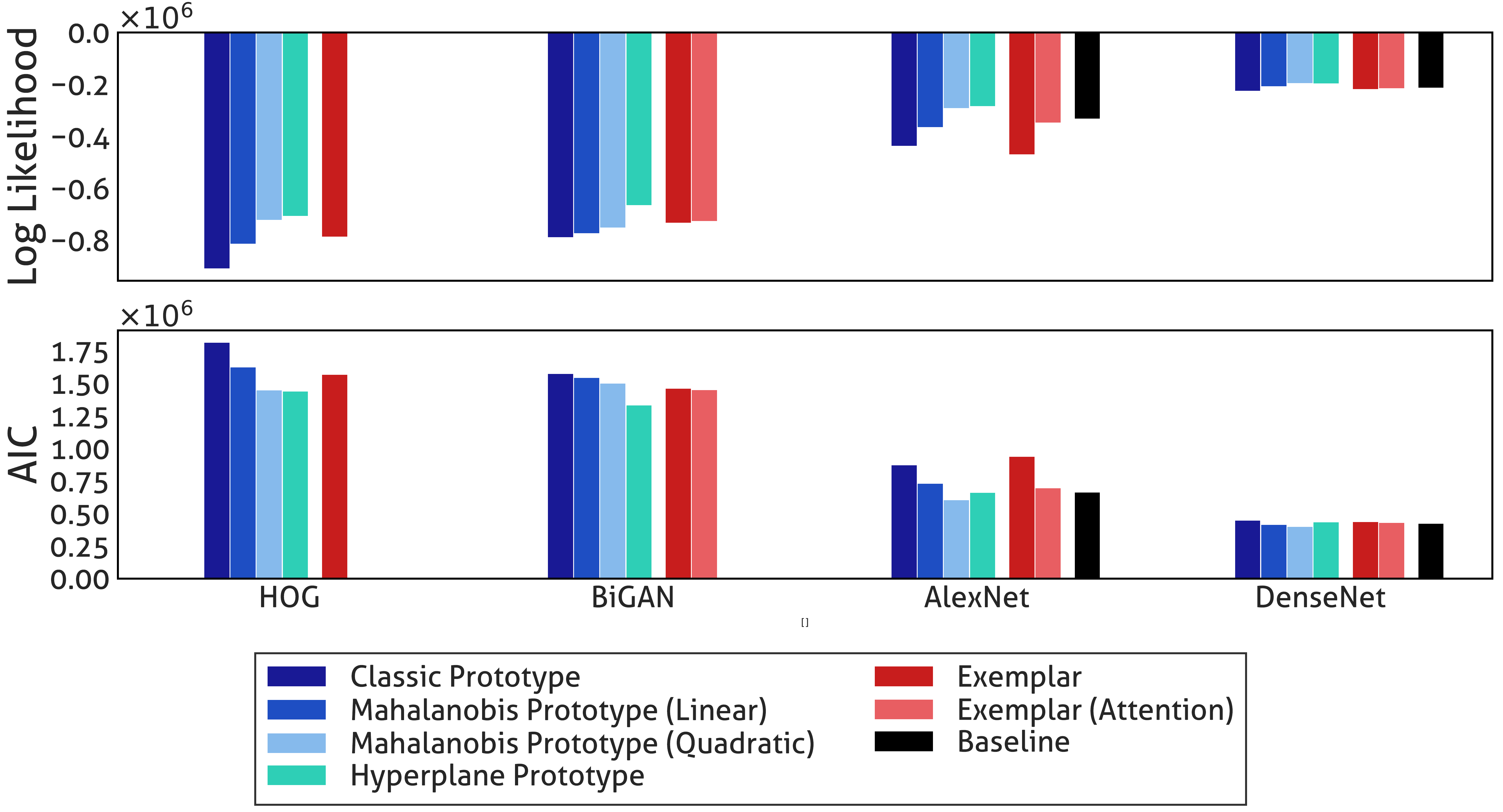}
	\caption{Log-likelihood and AIC results for all categorization models using all four stimulus representations. Within each representation, models are colored and clustered by class (prototype, exemplar). Baseline model performances are shown using a black bar for relevent feature spaces (AlexNet and DenseNet). For the HOG representations, the Exemplar (Attention) model failed to converge and is not shown.}
	\label{fig:results}
\end{figure}

Categorization model performance for each representation space is shown in Figure 4. One of the most salient overall outcomes is that the performance of all models is highly dependent on the underlying stimulus representation.
%
In particular, models making use of hand-engineered computer vision features (HOG) and those from an unsupervised generative CNN (BiGAN) exhibit poor fit to human behavior. In contrast, models perform much better when using representations from simple (AlexNet) and more advanced (DenseNet) supervised CNNs, achieving roughly half the (negative) log likelihood. This effect does appear to be obviously related to the dimensionality of stimulus representations: HOG and DenseNet representations are roughly the same dimensionality, and BiGAN representations comprise the smallest number of features. Nor is it explained by the use of convolutional architectures alone: BiGAN representations, while able to support effective generative image modeling and comparable downstream object classification using convolutional layers in the original work\cite{dumoulin2016adversarially}, perform considerably worse in modeling humans than even the much older supervised AlexNet CNN. Ultimately, models using the DenseNet representations performed best of all. This might be expected, given that these representations are from the CNN that scores highest on natural image ground truth classification. However, it is still interesting to note given that the DenseNet feature space is much more compressed than AlexNet---with roughly one third the number of dimensions---yet seems still to allow for more information relevant to the related task of human categorization to be retained.

As expected, both baseline models performed well. When using AlexNet and DenseNet representations, the following three models outperform the AlexNet and DenseNet baselines: the Hyperplane and Mahalanobis distance models from the linear prototype class and the Mahalanobis distance model from the quadratic prototype class. The Exemplar model with attention weights also performs well, roughly similar to baseline. On average the prototype and exemplar models perform comparably well, and, interestingly, the best-performing model for each set of representations is from the prototype class. This result is surprising given the frequent superiority in fit by exemplar models in previous work, and the fact that prototype models only make a single distance comparison per category, whereas exemplar models make $|C|=1{,}000$ comparisons in our dataset.

With our Hyperplane and Mahalanobis-distance prototype models, we learn the category prototype and variance, respectively. We find that while using supervised CNN features these strategies give equivalent results. For the HOG and BiGAN features, learning the prototype locations (Hyperplane model) appears to improve inference compared to learning category variances. These better performing prototype models have more free parameters, which allow them greater freedom to alter CNN representations using human behavioral data during training to learn better decision bounds. The Classic prototype model with one parameter, on the other hand, performed less well, and consistently worse than our baselines. This model does not incorporate information about human behavior to alter the shape of its decision boundary during training, instead estimating this from the CNN representations alone.

As the log-likelihood improvements for our best-performing models over the one-parameter Classic prototype model, two-parameter Exemplar model, and CNN baselines are partly a product of increased expressivity, we present AIC scores as an alternate measure of model fit which penalizes more highly-parameterized models. The only notable change when we do so is that the Hyperplane, which have parameters for every pair of categories, are penalized such that they perform as well as our baselines. This is unsurprising, as the complexity of these models grows with the square of number of categories, and in this study we have ten.

\section*{Discussion}
It is testament to the importance of categorization in accounts of cognition that it remains a focus of research and discovery after so many years of investigation. Continuing this rich tradition, in the present study we have asked whether we might be able to extend theory matured in simplified laboratory settings to capture human behavior over stimuli more representative of the complex visual world we have evolved and learned within. To do so, we take a novel methodological approach. First, we use a large, diverse collection of natural images as stimuli. Second, we use state-of-the-art methods from computer vision to {\it estimate} the structure of these stimuli, which our models further {\it adapt} based on behavioral training data. This contrasts with the majority of previous work, in which a small number of {\it a priori}-identified features were manipulated to {\it define} and differentiate categories, and as a consequence were limited to simple artificial stimuli. Finally, we offset the modeling uncertainty these advances introduce by using a large behavioral dataset to more finely assess graded category membership over stimuli. Taken together, our results show that using representations derived from CNNs makes it possible to apply psychological models of categorization to complex naturalistic stimuli, and that the resultant models allow for precise predictions about complex human behavior. Our most general finding is that categorization models that incorporate supervised CNN representations predict human categorizations over natural images well---in particular, an exemplar model and prototype models that are able to modify CNN representations using information about human categorizations. This is theoretically interesting because it indicates there is enough latent information and flexibility in the existing CNN representations to harness for such a related task. Moreover, the success of the prototype models in this context re-motivates these classic cognitive models even after decades or near dormancy.

Working with these complex, naturalistic stimuli reveals a potentially more nuanced view of human categorization. The broad consensus from decades of laboratory studies using simple artificial stimuli was that people could learn complex category boundaries of a kind that could only be captured by an exemplar model\cite{mckinley1995investigations}. Extrapolating from these results, we might imagine that categorization should be thought of in terms of learning complex category boundaries in simple feature-based representations. Our results outline a different perspective. Whereas our exemplar model performs well, there are clearly some representations---for example, those of the DenseNet---in which the simple category boundaries defined by prototype models seem sufficient to capture human behavior. When thinking about humans, feature representations are likely to have been learned early on, through a slow, data-driven learning process. Given these considerations, one might expect psychological representations to reflect the natural world, such that categorization of natural stimuli is made as efficient and as simple as possible. On the other hand, artificial or unlikely stimuli may at times carve out awkward boundaries in these spaces, which perhaps underlies the success of exemplar models up to this point. Incorporating feature learning into studies of human categorization has been called for in the past\cite{schynsgt98}, and developing a deeper understanding of how these processes interact, especially in the context of complex natural stimuli, is an important next step towards more fully characterizing human categorization.

We view our approach as complementary to recent work training CNNs to reproduce and predict the MDS-coordinates of naturalistic images and using these as input for categorization models\cite{sanders2018using}. The key difference is the source of stimulus representations, their generality, and their cost to procure. MDS-coordinate predictions, if reasonably approximated, capture a great deal of information about the aspects of stimuli that are psychologically meaningful. However, in order to train networks to predict them, human similarity judgments between images must be collected---an expensive task, as the number of pairwise similarity judgments grows quadratically with the number of images. Additionally, there is little guarantee CNNs trained to predict MDS-solutions for small sets of images will generalize well to more varied stimulus sets. By contrast, CNNs are routinely trained to classify thousands and even millions of natural images with high accuracy, indicating that the information in their representations is somewhat consistent for a wide range of complex visual stimuli. Using these direct estimations of stimulus structure could be considered a less precise approximation of human mental representations; however, the already highly relevant information they encode is easily improved by the simple transformations implicit in almost all of our top-performing models, which only require a training set of categorization data that grows linearly with the number of images.

The use and adaptation of machine learning techniques and representations to extend psychological research into more naturalistic domains is a field in its infancy. However, as we simply cannot access human psychological representations for such complex naturalistic stimuli, the preliminary success of the approaches above is encouraging, and we are likely to see further benefits that track the progress of large image databases and improvements in deep network architectures. More broadly, these results highlight the potential of a new paradigm for psychological research that draws on the increasingly abundant datasets, machine learning tools, and behavioral data available online, rather than procuring them for individual experiments at heavy computational and experimental cost. 
Towards this aim, the large dataset we offer in this work can provide a direct bridge between frontline efforts in machine learning and ecologically valid cognitive modeling, the two of which we hope can continue to develop in synergy.

\newpage
\section*{Author Contributions}
T.L. Griffiths developed the study concept. All authors contributed to the study design. J.C. Peterson collected the data and created the figures. R.M. Battleday analyzed the data and drafted the manuscript. All authors edited the manuscript.

\section*{Data availability}
All behavioral data is available on request from the authors, and to be deposited into a public repository prior to publication.

\section*{Code availability}
All computer code was written in open-source packages, and is available on request from the authors.

\section*{Ethical oversight}
We have complied with all relevant ethical regulations for this work. Informed consent was collected from all subjects, and ethical approval given to the study protocol by the IRB under protocol name ``Cognitive Research Using Amazon Mechanical Turk (Expedited)", number 2015-05-7551.
\newpage
\bibliography{BattledayMain}

\end{document}